\documentclass[conference]{IEEEtran}
\IEEEoverridecommandlockouts

\def\BibTeX{{\rm B\kern-.05em{\sc i\kern-.025em b}\kern-.08em
    T\kern-.1667em\lower.7ex\hbox{E}\kern-.125emX}}
\usepackage[utf8]{inputenc}
\usepackage{balance}

\def\BibTeX{{\rm B\kern-.05em{\sc i\kern-.025em b}\kern-.08em
    T\kern-.1667em\lower.7ex\hbox{E}\kern-.125emX}}
\usepackage{cite}
\usepackage{amsthm}
\usepackage{multirow}
\usepackage[table,xcdraw]{xcolor}
\usepackage{algorithm,algpseudocode}
\usepackage{hyperref}
\usepackage{url}
\usepackage{xspace}
\usepackage{graphicx}
\usepackage{amsfonts}
\usepackage{amsmath,amssymb,amstext}
\usepackage{textcomp}
\usepackage{acro}
\usepackage{enumitem}
\usepackage{pgfplots}
\usepackage{pgfplotstable}
\pgfplotsset{compat=newest}
\usepgfplotslibrary{groupplots,dateplot}
\usepackage{soul}
\usepackage{mathtools,lipsum, nccmath,cuted,amsmath}
\usepackage[capitalise]{cleveref}
\usepackage{siunitx}
\usepackage{textcomp}

\usepackage{comment}
\usepackage{makecell}
\usepackage{algorithm}
\usepackage{gensymb}
\usepackage[normalem]{ulem}
\usepackage{cancel}
\usepackage{subcaption}
\DeclareMathOperator*{\argmax}{argmax}
\newtheorem*{problem}{Problem}
\newtheorem*{example}{Example}
\let\oldexample\example
\renewcommand{\example}{\oldexample\normalfont}
\usepackage{listings}%
\definecolor{keywords}{HTML}{8A4A0B}

\definecolor{background}{HTML}{EEEEEE}

\definecolor{comments}{HTML}{868686}

\lstdefinelanguage{sdl}{
 morekeywords={scene,long,type,entity,'{','}', class},
 keywordstyle=\color{keywords},
    basicstyle=\scriptsize\ttfamily,
 morecomment=[l]{//}, 
 morecomment=[s]{/*}{*/}, 
 morestring=[b]",
    basicstyle=\scriptsize\ttfamily,%
 commentstyle=\color{comments}\ttfamily,
numbers=right,
    numberstyle=\scriptsize,
    stepnumber=1,
    numbersep=8pt,
breaklines=true,
    frame=tb,
 tabsize=4}

\usetikzlibrary{pgfplots.statistics, pgfplots.colorbrewer,pgfplots.dateplot,
        shadows,
        matrix} 

\definecolor{blueLine}{RGB}{57,106,177}
\definecolor{blueFill}{RGB}{114,147,203}
\definecolor{redLine}{RGB}{204,37,41}
\definecolor{greenLine}{RGB}{0,250,0}
\definecolor{blackLine}{RGB}{0,0,0}
\definecolor{goldLine}{RGB}{160,82,45}
\definecolor{brightgreen}{rgb}{0.4, 1.0, 0.0}
\definecolor{brinkpink}{rgb}{0.98, 0.38, 0.5}
\definecolor{cadmiumyellow}{rgb}{1.0, 0.96, 0.0}
\definecolor{cinnamon}{rgb}{0.82, 0.41, 0.12}
\definecolor{darkorange}{rgb}{1.0, 0.55, 0.0}
\definecolor{darkspringgreen}{rgb}{0.09, 0.45, 0.27}

\usepackage[font=footnotesize]{caption}
\usepackage{soul}
\usepackage{xcolor}
\usepackage{multirow}
\usepackage{soul}

\author{\IEEEauthorblockN{Shreyas Ramakrishna, Baiting Luo, Yogesh Barve, Gabor Karsai, and Abhishek Dubey}
\IEEEauthorblockA{Institute for Software Integrated Systems, Vanderbilt University}\\
[-4.5ex]}

\DeclareAcronym{tfpg}{
  short = TFPG,
  long  = Timed Failure Propagation Graph,
}
\DeclareAcronym{ac}{
  short = AC,
  long  = Assurance Case,
}

\DeclareAcronym{bo}{
  short = BO,
  long  = Bayesian Optimization,
}

\DeclareAcronym{dnn}{
  short = DNN,
  long  = Deep Neural Network,
}

\DeclareAcronym{gbo}{
  short = GBO,
  long  = Guided Bayesian Optimization,
}

\DeclareAcronym{rns}{
  short = RNS,
  long  = Random Neighborhood Search,
}

\DeclareAcronym{cps}{
  short = CPS,
  long = Cyber Physical System,
}

\DeclareAcronym{aebs}{
  short = AEBS,
  long = Automatic Emergency Braking System,
}
\DeclareAcronym{sdl}{
  short = SDL,
  long = Scenario Description Language
}

\DeclareAcronym{av}{
  short = AV,
  long = Autonomous Vehicle
}

\DeclareAcronym{resonate}{
  short = ReSonAte,
  long  = Runtime Safety Evaluation in Autonomous Systems,
}

\DeclareAcronym{dsml}{
  short = DSML,
  long  = Domain-Specific Modeling Language,
}
\DeclareAcronym{lec}{
  short = LEC,
  long = Learning Enabled Component,
}

\DeclareAcronym{cnn}{
  short = CNN,
  long = Convolutional Neural Network,
}

\DeclareAcronym{btd}{
  short = BTD,
  long  = Bow-Tie Diagram,
}

\DeclareAcronym{ood}{
  short = OOD,
  long = Out-of-Distribution
}

\DeclareAcronym{vae}{
  short = VAE,
  long = Variational Autoencoder,
}

\DeclareAcronym{gan}{
  short = GAN,
  long = Generative Adversarial Network,
}

\DeclareAcronym{gps}{
  short = GPS,
  long = Global Positioning System
}
\DeclareAcronym{imu}{
  short = IMU,
  long = Inertial Measurement Unit
}

\DeclareAcronym{rl}{
  short = RL,
  long = Reinforcement Learning
}

\DeclareAcronym{ucb}{
  short = UCB,
  long = Upper Confidence Bound
}
\DeclareAcronym{lut}{
  short = LUT,
  long = Lookup Table
}

\DeclareAcronym{gp}{
  short = GP,
  long = Gaussian Process
}

\usepackage{todonotes}

\begin{document}

\title{Risk-Aware Scene Sampling for Dynamic Assurance of Autonomous Systems}






\maketitle
\setcounter{page}{1}
\begin{abstract}
Autonomous Cyber-Physical Systems must often operate under uncertainties like sensor degradation and shifts in the operating conditions, which increases its operational risk. Dynamic Assurance of these systems requires designing runtime safety components like Out-of-Distribution detectors and risk estimators, which require labeled data from different operating modes of the system that belong to scenes with adverse operating conditions, sensors, and actuator faults. Collecting real-world data of these scenes can be expensive and sometimes not feasible. So, scenario description languages with samplers like random and grid search are available to generate synthetic data from simulators, replicating these real-world scenes. However, we point out three limitations in using these conventional samplers. First, they are passive samplers, which do not use the feedback of previous results in the sampling process. Second, the variables to be sampled may have constraints that are often not included. Third, they do not balance the tradeoff between exploration and exploitation, which we hypothesize is necessary for better search space coverage. We present a scene generation approach with two samplers called Random Neighborhood Search (RNS) and Guided Bayesian Optimization (GBO), which extend the conventional random search and Bayesian Optimization search to include the limitations. Also, to facilitate the samplers, we use a risk-based metric that evaluates how risky the scene was for the system. We demonstrate our approach using an Autonomous Vehicle example in CARLA simulation. To evaluate our samplers, we compared them against the baselines of random search, grid search, and Halton sequence search. Our samplers of RNS and GBO sampled a higher percentage of high-risk scenes of 83\% and 92\%, compared to 56\% 66\% and 71\% of the grid, random and Halton samplers, respectively.

\end{abstract}

\begin{IEEEkeywords}
Cyber-Physical Systems, Dynamic Assurance, Scenario Description Language, Bow-Tie Diagram
\end{IEEEkeywords}
\pagestyle{plain}
\section{Introduction}
The widespread use of autonomous \acp{cps}\footnote{CPS with \ac{lec}} often requires them to operate under uncertainties like sensor degradation and shifts in the operating conditions, which increase its operational risk. Design-time Assurance Case~\cite{bishop2000methodology} with risk assessment information is used to argue the system's safety at runtime. However, the dynamically changing operating conditions of the system at runtime potentially invalidate the design-time assumptions and the safety arguments~\cite{denney2015dynamic}. A dynamic approach with proactive safety assessment components like \ac{ood} detectors~\cite{sundar2020out,ramakrishna2021efficient}, failure predictors~\cite{kuhn2020introspective,kuhn2021trajectory} and dynamic assurance monitors~\cite{hartsell2021resonate} is required for runtime safety assurance. Designing these components often requires labeled data from different operating modes of the system that belong to scenes with adverse operating conditions and sensor or actuator faults. These scenes are referred to as risky scenes~\cite{ma1999worst} or safety-critical scenarios~\cite{ding2020learning}; in this paper, we refer to them as high-risk scenes.

Often, the data related to high-risk scenes are under-represented in the training sets~\cite{khan2019striking}, leading to a data imbalance problem. If we can generate these under-represented events, they can be used to design the safety assessment components required for dynamic assurance and retrain the controller LECs to improve their accuracy~\cite{fremont2019scenic}. However, collecting real-world data of such high-risk scenes can be expensive and slow in real-world conditions. Synthetic data from simulators have been used to address this problem in engineering design and testing applications. Samplers are used to generate data across the search space created by the system parameters. For example, tools like Dakota~\cite{dalbey2021dakota} provide efficient samplers like incremental sampling, importance sampling, and adaptive sampling for uncertainty quantification in engineering design. Recently, this concept of sampling-based data generation is being adapted for autonomous systems~\cite{zhao2017accelerated,ma1999worst,dreossi2019verifai,viswanadha2021parallel}. Domain-specific \ac{sdl} like Scenic \cite{fremont2019scenic}, and MSDL \cite{msdl} with 
conventional samplers like random and grid search are integrated with simulators like CARLA \cite{dosovitskiy2017carla} to generate high-risk scenes. 


Despite their success in generating high-risk scenes, we point to several limitations in using the conventional samplers. First, they perform \textit{passive sampling}, which does not use the feedback of previous results in the sampling process. Second, the scene variables (e.g., environmental conditions) being sampled typically have \textit{sampling constraints} and \textit{co-relations} that need to be considered. For example, environmental conditions (e.g., precipitation) may have physical constraints on their values that govern their temporal evolution. Applying these constraints is necessary for generating meaningful scenes~\cite{fremont2019scenic}. However, the conventional samplers do not include these sampling constraints. Third, conventional samplers do not balance the \textit{exploration vs. exploitation trade-off}. For example, random and Halton searches prioritize uniform search space coverage, so they only explore. In contrast, grid search aims to cover a given grid exhaustively, so they only exploit it. However, as discussed by Jerebic, Jernej \emph{et al.}~\cite{jerebic2021novel}, balancing these strategies can result in higher coverage diversity which is commonly measured using clustering properties like the number of clusters and cluster population. 

To address these limitations, we present a scene generation approach that has a domain-specific \ac{sdl} integrated with two sampling approaches for generating high-risk simulation scenes. The key contribution of this work is two sampling approaches called \ac{rns} and \ac{gbo}, which perform active sampling by using previous simulation results in the sampling process. They also include sampling constraints that govern the evolution of scene variables. In addition, they provide explicit hyperparameters to control the tradeoff between exploration vs. exploitation of the search strategy. Further, to facilitate the samplers, we use a novel risk-based scoring function that evaluates how risky the scene was for the system. We demonstrate our approach using an \ac{av} case study in the CARLA simulator \cite{dosovitskiy2017carla}. To evaluate our samplers, we compared them against the baselines of random search, grid search, and Halton sequence search~\cite{halton1960efficiency} in terms of three metrics that measure the total high-risk scenes sampled, the sample diversity, and the search times. Our experimental evaluation shows that the \ac{rns} and \ac{gbo} outperform the baselines in terms of the total high-risk scenes and diversity metrics. Also, the \ac{rns} sampler had comparable search times to the baselines, but the \ac{gbo} sampler has a higher search time. The source code is available online\footnote{\label{github}https://github.com/scope-lab-vu/Risk-Aware-Scene-Generation-CPS}.

\begin{figure}[t]
 \centering
 \includegraphics[width=0.9\columnwidth]{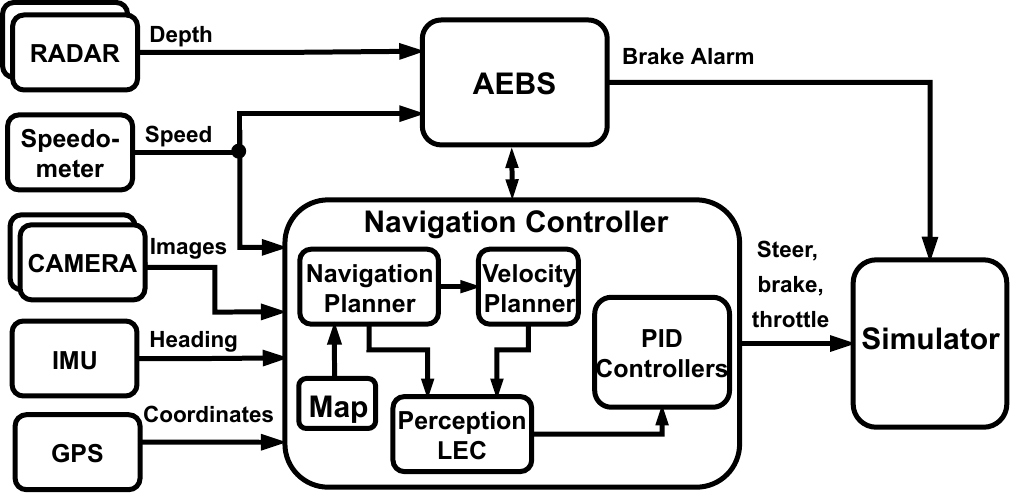}
 \caption{\ac{av} system model designed for the CARLA autonomous challenge~\cite{carla-challenge} setup. The \ac{av} is primarily driven by an \ac{lec}-based navigation controller adapted from Chen et al.~\cite{chen2020learning}. We have augmented an \ac{aebs} supervisor for emergency braking.}
 \label{fig:sys-model}
\vspace{-0.1in}
\end{figure}

The outline of this paper is as follows. In \cref{sec:background}, we introduce the background needed to understand this work. In \cref{sec:ps}, we describe the problem statement, followed by a description of our approach in \cref{sec:approach}. In \cref{sec:evaluation}, we implement and evaluate the samplers in the context of an \ac{av} case study in the CARLA autonomous driving challenge~\cite{carla-challenge}. This setup requires an \ac{av} (See \cref{fig:sys-model}) to navigate an urban town setting with complex traffic scenarios, adverse weather conditions, and sensor faults. Finally, we present related research in \cref{sec:rw} followed by conclusions in \cref{sec:conclusion}.

\begin{figure*}[t]
 \centering
 \includegraphics[width=\textwidth]{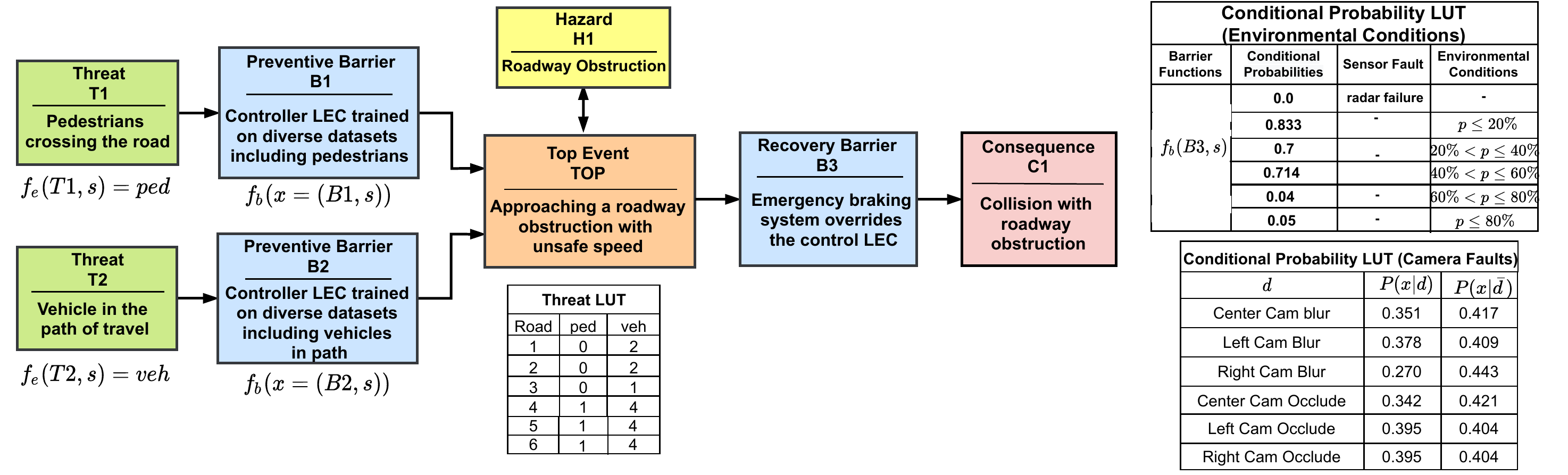}
 \caption{Bow-Tie Diagram for the \ac{av} case study adopted from our previous work~\cite{hartsell2021resonate}. Each block includes information about the event type and its description. The conditional probability and the threat $LUT$s are estimated from calibration data, which is discussed in \cref{sec:risk}.}
 \label{fig:btd}
\end{figure*}
\section{Background}
\label{sec:background}


\subsection{System Design Procedures}
The typical design procedure of autonomous \acp{cps} include the design, training, testing, and deployment phases, which we categorize into five steps for designing our systems: (1) \textit{Design Phase}, which involves system analysis, hazard analysis, and Assurance Case construction. (2) \textit{Training Phase}, which involves collecting training scenes and training the \acp{lec} on these scenes. (3) \textit{Calibration Phase}, which involves calibrating the detectors~\cite{cai2020real,sundar2020out}, and dynamic assurance monitors (\cref{sec:risk_estimation}). Calibration requires curating a \textit{calibration set} that includes scenes with sensor faults and adverse weather in addition to the training scenes. To generate these scenes, we use both random and grid samplers. We use the random sampler to find the conditions affecting the system and then use the grid sampler to generate more scenes around it. (4) \textit{Testing Phase}, which involves generating high-risk scenes for testing the trained system. (5) \textit{Deployment Phase}, which involves deploying the trained and tested system to operate. 



In particular, the design phase is important because it consists of gathering and documenting information about the system's goals, requirements, operating conditions, and component faults. It also includes designing the system models like the architecture model (e.g., \cref{fig:sys-model}) and the system function breakdown model. Following this, the designers perform an \textit{hazard analysis}, which involves identifying the hazards to the system that will result in a system consequence. The system could have a collection of hazards $\{H_1, H_2,\cdots, H_n\}$ stemming from its operation, functioning, software components, and hardware components. For example, in the context of an \ac{av}, operational hazard elements can be the movement of traffic participants including pedestrians, and other vehicles. Any of these hazards in a given condition can result in a system consequence. This enables the creation of \ac{btd}~\cite{ferdous2013analyzing}, a model describing the chain of events $t_i \rightarrow b_p \rightarrow e_{top} \rightarrow b_m \rightarrow c_i$, where $t_i$ are system threats, $b_p$ are preventive barriers, $e_{top}$ is a top event, $b_m$ is a mitigation barrier, and $c_i$ is a system consequence. \cref{fig:btd} describes the operational hazard of ``roadway obstruction'' for the \ac{av} case study. The \ac{btd} has two system threats: vehicles (T1) and pedestrians (T2) in the path of the ego vehicle. These threats can escalate to become a top event (TOP) if not prevented by the preventive barriers $B1$ and $B2$. Next, the TOP event can escalate to become a system consequence if not mitigated by the barrier $B3$. During the design, these hazard analyses are used to create assurance cases~\cite{bishop2000methodology}, showing that the top-level goals of the system are satisfied. Further, the \ac{btd} allows for dynamically monitoring the likelihood of system consequences through a technique called \textit{dynamic assurance}. 

\subsection{ReSonAte}
\label{sec:risk_estimation}

To perform dynamic assurance, we leverage our previously developed tool called \ac{resonate} ~\cite{hartsell2021resonate}, which uses an augmented \ac{btd} derived from the hazard analysis. The augmentation adds various event probabilities for each causal chain, conditioned on the state of the system and the environment, including sensor failures, actuator failures, the output of anomaly detectors, and environmental conditions. Specifically, we need to estimate the (a) the conditional probability of the barrier's success in each state $f_b(x=(B_i,s))$, and (b) the frequency of occurrence of a threat $f_e(T_i,s)$ in each state. We infer these conditional relationships from the calibration dataset gathered during the previously discussed calibration phase. The estimated probabilities are stored in \acp{lut} and used at runtime to calculate the hazard rate $\lambda$ and the likelihood of the hazard occurrence in each time unit (t) as $1-e^{-\lambda \cdot t}$.

\section{Problem Formulation}
\label{sec:ps}
We consider the autonomous \ac{cps} to be operating in an environment characterized by two sets of variables $\mathbb{E}: x \rightarrow \Re^{+}$ and $\mathbb{S}: y \rightarrow \Re \times \Re$. $\mathbb{E}$ are the environmental variables like rain, traffic density, time-of-the-day. $\mathbb{S}$ are structural variables related to roadway features and are characterized by waypoints $w$ denoted by a two-dimensional matrix of latitude and longitude. Also, we can map each waypoint to a road segment. In addition, we consider the set of faults in the system sensors and actuators $F: x \rightarrow {0,1}$. For a particular operating environment, there can be several environment variables $e \in \mathbb{E}$, several possible waypoints $w \in \mathbb{S}$, and several sensors in the system that can fail $f \in \mathbb{F}$. The \ac{cps} is trained through a collection of scenes, where each scene $s_i$ is an ordered sequence of $k$ sampled observations $< \mathbb{E}_i, \mathbb{S}_i, \mathbb{F}_i >_{i=1}^{k}$. We collectively refer to these variables as the scene variables $s_v$. Given each scene, we can associate the hazard rate $\lambda$ for every hazard identified for the system. The system can have a collection of hazards $\{H_1,H_2,\cdots,H_n\}$ stemming from its components (e.g., software, hardware) or its operating conditions. Each identified hazard will have an associated hazard rate estimated using the dynamic assurance routines ($DA: s_i \rightarrow \lambda$) introduced in \cref{sec:risk_estimation}. 
We can use the estimated hazard rate to compute the likelihood of the hazard occurrence ($1-e^{-\lambda \cdot t}$) in each time unit $t$. The likelihood of hazard for different hazard conditions put together constitutes the system's operational risk $S_{Risk}$.


\begin{table}[!t]
\centering
\renewcommand{\arraystretch}{1.1}
\footnotesize
\begin{tabular}{|c|c|c|c|}
\hline
\textbf{Variable Type}                   & \textbf{Name}                                                   & \textbf{Range} & \textbf{Constraints}                                                                \\ \hline
\textbf{Structural}                     & \begin{tabular}[c]{@{}c@{}}Road Segments\\ (RS)\end{tabular}    & {[}0,9{]}      & $wp_1 - wp_2 \leq 10m$                                      \\ \hline
\multirow{4}{*}{\textbf{Environmental}} & Precipitation (P)                                               & {[}0,100{]}    & $P_i - P_{i+1} \leq 5\%$                                                            \\ \cline{2-4} 
                                        & Time-of-Day (T)                                                 & {[}0,90{]}     & $T_i - T_{i+1} \leq 10^{\circ}$                                                            \\ \cline{2-4} 
                                        & Cloud (C)                                                       & {[}0,100{]}    & $C_i - C_{i+1} \leq 5\%$                                                            \\ \cline{2-4} 
                                        & \begin{tabular}[c]{@{}c@{}}Traffic Density \\ (TD)\end{tabular} & {[}0,20{]}     & \begin{tabular}[c]{@{}c@{}}$TD_i$ depends on $RS_i$\\ $TD_i - TD_{i+1} \leq 10$\end{tabular} \\ \hline
\multirow{2}{*}{\textbf{Faults}}        & Cam Blur                                                 & {[}0,1{]}      & \multirow{2}{*}{N/A}                                                                \\ \cline{2-3}
                                        & \begin{tabular}[c]{@{}c@{}}Cam Occlusion \end{tabular} & {[}0,1{]}      &                                                                                     \\ \hline
\end{tabular}
\caption{Scene variables, their distribution ranges, and sampling constraints for the CARLA \ac{av} case study introduced later in \cref{sec:evaluation}.} 

\label{Table:constraints}
\vspace{-0.1in}
\end{table}


\begin{figure*}[t]
\centering
 \includegraphics[width=\textwidth]{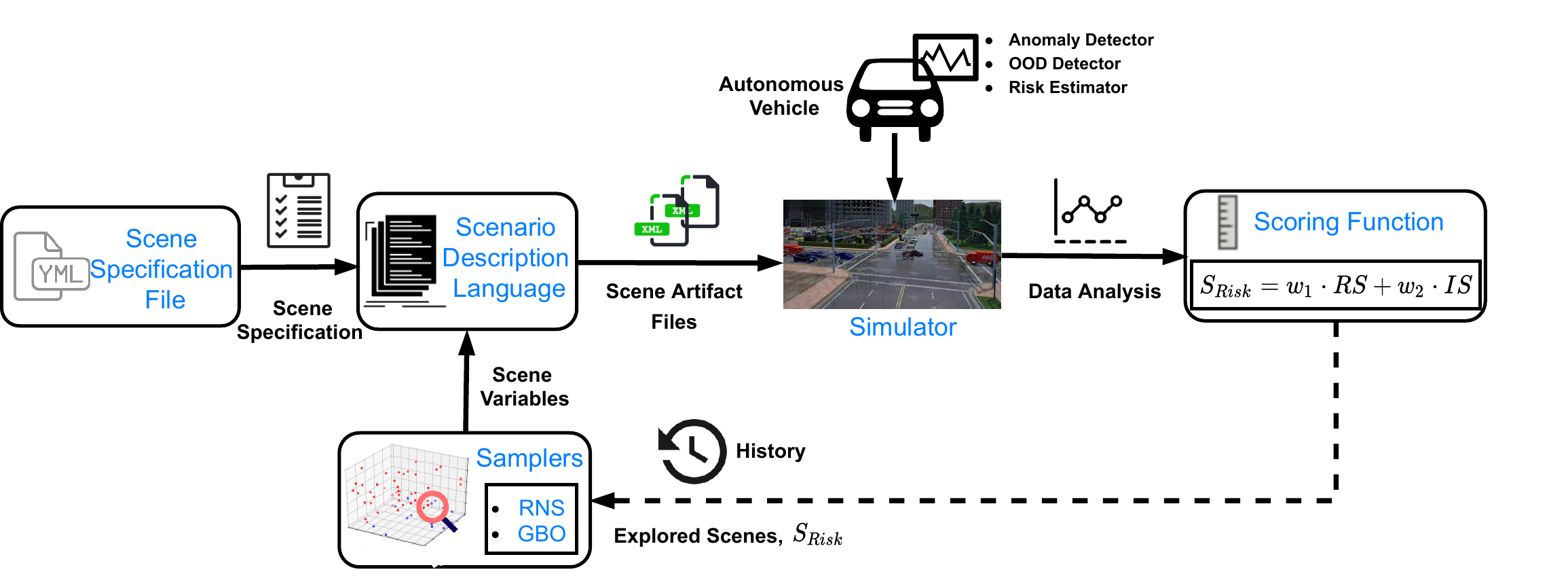}
 \caption{Overview of our approach for generating high-risk scenes. The samplers perform active sampling using the feedback loop represented by the dotted line.}
 \label{fig:workflow}
\end{figure*}


Also, given a sequence of samples in the scenes, physical constraints govern the temporal evolution of the sampled values. For example, a constraint on an environmental variable $e \in \mathbb{E}$ governs that $|s_i^e-s_{i+1}^e| \leq \alpha_e \in \Re^+$. Similarly, the waypoints across a scene are governed by the routability property. To explain this, we use the sampling constraints $\mathcal{SC}$ in \cref{Table:constraints}, which we applied for an \ac{av} case study, discussed in \cref{sec:casestudy}. The constraint governs the road segment such that the distance between any two selected waypoints cannot exceed $10$ meters. This constraint regulates that the road segment is orderly selected. The environmental variables have constraints on how their values change between consecutive scenes. The traffic density has constraints on its value and is also dependent on the type of road segment (e.g., crossroads, intersections). Also, we do not place any constraints on the fault variables. Now, given this setup, we want to address the following problem. 


\begin{problem}
\label{prob:high-risk scene}
Given a threshold $\delta$ (computed across the calibration set of scenes), sample $N$ scenes that maximizes the risk $\max(sum(S_{Risk} - \delta))$. In addition, we want to ensure that the scene variables are diverse. We call a scene with ($S_{Risk} - \delta) > 0$  as a high risk scene.


\end{problem} 

While the first objective measures the sampler's capability to sample high-risk scenes, the second measures its coverage diversity, representing its capability to balance exploration vs. exploitation of the search space. A sampler with high diversity is desirable for better coverage of the search space~\cite{jerebic2021novel}. We measure diversity in terms of a quantitative metric (measured using the number of the optimal clusters, as measured using silhouette score~\cite{rousseeuw1987silhouettes}) and a qualitative metric (measured using the variance of risk across each scene across all clusters).

\section{Proposed Approach}
\label{sec:approach}

We now describe the approach we use to generate high-risk scenes for \ac{av} in simulation. An overview of the approach is illustrated in \cref{fig:workflow} and described in detail below. 

\subsection{Scene Generation}
\label{sec:scene_generation}
We have designed a scene generation approach, which includes a \ac{sdl} integrated with samplers to sample scenes over different scene variables introduced in \cref{sec:ps}. The \ac{sdl} is designed using the textX \cite{dejanovic2017textx} meta-language and it has two components. (a) \textit{grammar}, which consists of a set of rules required to define a scene in the meta-language. (b) \textit{meta-model}, which defines the concept of a scene using scene variables and their distribution ranges. In this setup, a scene $s$ = \{$e_1$,$e_2$,...,$e_k$\} is a collection of entities, where each entity is a tuple $<s_v,v_d,v_c>$ of the scene variables $s_v$, distribution properties $v_d$ (e.g., distribution type, ranges) and variable constraints $v_c$. The \ac{sdl} also has an interpreter to generate the scene artifact files that run the simulator. The interpreter also abstracts the complexity of the \ac{sdl} from the user by providing a scene specification file with scene variables and samplers that can be chosen to generate different scenes. A fragment of the scene description for CARLA simulation is shown in \cref{fig:scene-meta}.

 

\begin{figure}[!t]
\setlength{\abovecaptionskip}{-3pt}
\begin{lstlisting}[language=sdl,numbers=none]
scene sample {
type int
type string
class scene_info {
    town: town_description
    weather: weather_description
    road: road_segments
    traffic: traffic_density
    faults: sensor_faults }
entity town_description{
        id:string
        map:string  }
entity weather_description {
    precipitation: distribution
    cloudiness: distribution
    time-of-day: distribution }
entity distribution {
        low: int
        high: int  }
}
 
\end{lstlisting}
\caption{This listing shows a fragment of the scene description for CARLA simulation that was generated using our \ac{sdl}.}
\label{fig:scene-meta}
 \vspace{-0.15in}
\end{figure}


\subsection{Scoring Function}
\label{sec:scoring}

We have designed the \textit{Risk Score} metric ($S_{Risk}$) to evaluate the sampled scene. This score is a weighted combination of two utility functions and is computed as $S_{Risk} =  w_1 \cdot RS + w_2 \cdot IS$. The utility functions are. (1) \textit{\ac{resonate} Score} ($RS$), which measures the system's risk in an operational scene using the hazard rates calculated by the \ac{resonate} framework. (2) \textit{Infraction Score} ($IS$), which measures the actual infractions performed by the system. We use equal weights of $w_1$ = $w_2$ = 1 for both these utility functions to ensure that the estimated risk and the actual risk caused by infractions are weighted equally in the subsequently sampled scenes. In the future, we will formulate this metric as a multi-objective optimization problem as performed by Viswanadha, Kesav, \emph{et al.}~\cite{viswanadha2021parallel}.








\subsubsection{\ac{resonate} Score ($RS$)} Measures the average risk of the system's failure across a scene. We leverage the \ac{resonate} framework which estimates the hazard rate $\lambda$, which is averaged across the scene using $\frac{\int_{T_1}^{T_2} \lambda \,dt}{T_2-T_1}$ to compute the system's risk or the \ac{resonate} score. $\lambda$ is the estimated hazard rate and $T_1$, $T_2$ are the start and end time of the scene.


\subsubsection{Infraction Score ($IS$)} Measures the system's actual infractions across a scene. This score can include a variety of infractions like route deviation, lane violation, and traffic rule violation. These infractions ($I_k$) can also be weighted ($w_k$) according to their severity and later summed together to compute a unified infraction score $IS$ = $\sum_{k=1}^{n} w_k \cdot I_k$. To illustrate, the infraction score for the \ac{av} case study was computed as $IS = 0.7 \cdot I_S + 0.8 \cdot I_R + I_{RD}$. $I_S$ is running a stop sign infraction, $I_R$ is running a red-light infraction, and $I_{RD}$ is the deviation in the route taken by the \ac{av}. We adopted these weights from the CARLA challenge setup~\cite{carla-challenge}.





Assuming these scores are available at the end of each scene, we compute the risk score $S_{Risk}$. Further, a scene is classified to be of high-risk if $S_{Risk} - \delta$ $>$ 0. $\delta$ is the risk threshold computed across the previously encountered calibration scenes. Considering these scenes are sampled from the same underlying distribution, we select the threshold at the $95^{th}$ percentile of $S_{Risk}$ for every scene in the calibration set. 



\begin{algorithm}[!t]
    \footnotesize{}
    \caption{Random Neighborhood Search}
	\textbf{Parameter}: number of iterations $t$, explored list $\mathcal{E}$, neighborhood size $k$\\
	\textbf{Input}: search space $\mathcal{D}$, sampling constraints $\mathcal{SC}$, scene variables $s_v$, threshold $\delta$\\
	\textbf{Output}: List of scenes
	\begin{algorithmic}[1]
	\For{$x=1,2,...,t$}
	   \If{$S_{Risk} < \delta$ or ($S_{Risk} > \delta$ and $N \geq k$)}
	        \State Randomly sample $s_v$ within $\mathcal{D}$ to generate random scene $R$
	    \Else
	        \State Apply $\mathcal{SC}$ to create bounded search area $\mathcal{B}$ 
	        \State Randomly sample $s_v$ within $\mathcal{B}$ to generate neighboring scene $N$
	        \State Apply kd-tree to find if there are atleast $k$ neighbors ($N$) in $\mathcal{E}$
	    \EndIf
        \State Use the sampled variables to generate the scene artifact files
        \State Simulate the scene and compute the $S_{Risk}$ 
        \State Append sampled variables and $S_{Risk}$ to $\mathcal{E}$ 
	\EndFor
	\end{algorithmic} 
	\normalsize{}
\label{algo:rns}
\end{algorithm}

\subsection{Samplers}
\label{sec:samplers}
We have developed two samplers called the \acl{rns} and \acl{gbo}, which are extensions of the conventional random and \ac{bo} search. The extensions are. (1) \textit{active sampling}, we use a feedback loop of previous results to sample the next scene variables. (2) \textit{constraints-based search}, we create sampling constraints $\mathcal{SC}$ to add constraints on the scene variables. \cref{Table:constraints} shows the constraints that we applied for an \ac{av} case study. (3) \textit{exploration vs. exploitation trade-off}, we introduce explicit hyper-parameters to balance the strategies.

\subsubsection{Random Neighborhood Sampler (RNS)}
This sampler extends the conventional random search by including the kd-tree nearest neighborhood search algorithm~\cite{friedman1975algorithm}. We aim to add the missing exploitation capability to random search with this extension. Briefly, the approach works as follows. It initially explores a scene through random sampling and then exploits the area around the explored scene using the sampling constraints and the nearest neighbor search. \cref{algo:rns} illustrates the steps involved, and it works as follows. 

\textit{First}, the algorithm explores the search space $\mathcal{D}$ by randomly sampling the scene variables $s_v$ from their respective distribution range $v_d$. These randomly selected variables are used to generate the scene artifact files, run the simulator and compute the risk score $S_{Risk}$ for the scene. 


\textit{Second}, if the $S_{Risk}$ $<$ $\delta$, the statistics of the scene are stored, and the scene variables are re-sampled again from their respective distribution range $v_d$. However, if the $S_{Risk}$ $>$ $\delta$, the neighborhood around the randomly sampled scene $R$ is exploited to generate more scenes. For this, we create a bounded region $\mathcal{B}$ around $R$ by applying the sampling constraints $\mathcal{SC}$ to the distribution ranges of the scene variables.


\textit{Third}, the algorithm uses the kd-tree nearest neighborhood search to find if there are at least $k$ scenes generated in the neighborhood of the explored scene $R$. That is, it checks if $R$ has at least $k$ similar scenes in $\mathcal{E}$. To measure the similarity, we use the $l_2$ distance metric $(d(x,R) = ||x - R||_2)$ and identify the similar scenes as $N = \{\forall s_l \in \mathcal{E}| d(R,s_l) < \tau \}$. Where $R$ is the currently sampled scene, and $s_l$ are previously explored scenes in the explored list $\mathcal{E}$. This step returns a list of neighbors whose distances to $R$ are smaller than a threshold $\tau$.  If the number of the neighbors $N$ $\geq$ $k$, the search around $R$ is stopped. Then, the first step is repeated to randomly explore a new scene in the entire search space $\mathcal{D}$. However, if $N$ $\leq$ $k$, then the search around $R$ is continued. In this sampler, $k$ is the hyperparameter used to control the exploitation.



\begin{algorithm}[!t]
    \footnotesize{}
    \caption{Guided Bayesian Optimization}
	\textbf{Parameter}: number of iterations $t$, initial iterations $k$, explored list $\mathcal{E}$\\
    \textbf{Input}: search space $\mathcal{D}$, sampling constraints $\mathcal{SC}$, scene variables $s_v$, threshold $\delta$\\
	\textbf{Output}: List of scenes 
	\begin{algorithmic}[1]
	\For{$x=1,2,...,t$}
	    \If{$x \leq k$}
	        \State initialize GP model with random samples $s_v$ from $\mathcal{D}$
	    \Else
	        \State Apply $\mathcal{SC}$ to create bounded search area $\mathcal{B}$
	         \State Use $\mu_t$ and $\sigma_t$ in the UCB function to sample $s_v$ within $\mathcal{B}$
	    \EndIf
        \State Use the sampled variables to generate the scene artifact files
        \State Simulate the scene and compute the $S_{Risk}$ 
        \State Append sampled variables and $S_{Risk}$ to $\mathcal{E}$ 
        \State Update the GP model using $\mathcal{E}$. Update $\mu_t$ and $\sigma_t$
	\EndFor
	\end{algorithmic} 
	\normalsize{}
\label{algo:bo}
\end{algorithm}

\begin{figure*}[t]
 \centering
 \includegraphics[width=0.84\textwidth]{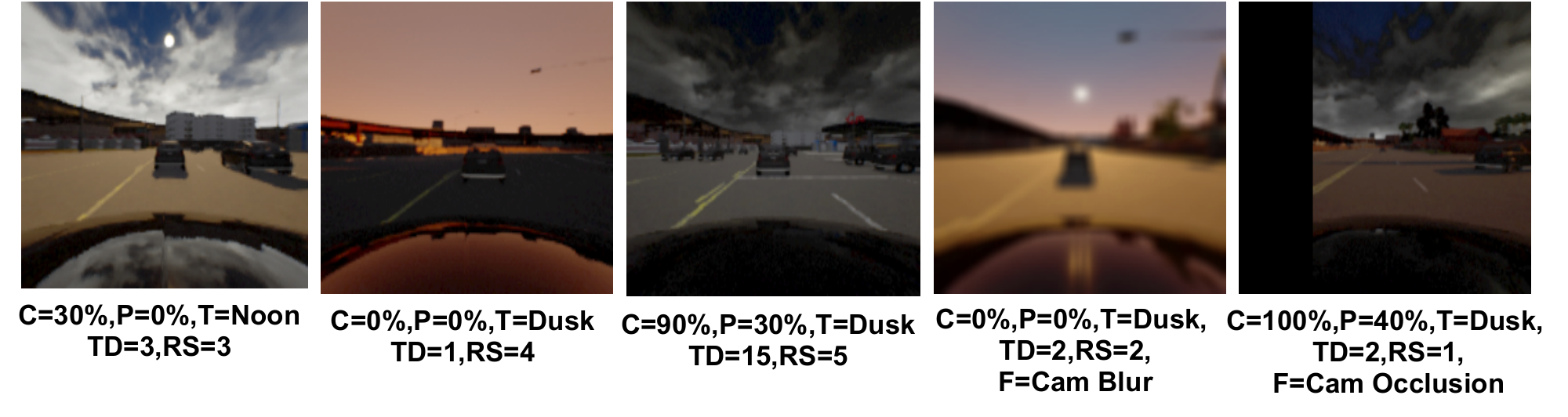}
 \caption{Screenshots of different scenes generated for the CARLA \ac{av} case study. The description of these scenes are provided below the images.}
 \label{fig:scenes}
\vspace{-0.1in}
\end{figure*}

Although the sampler can both explore and exploit, the exploration is performed uninformed without strategic knowledge about the search space. This could result in high-risk regions of the search space left unexplored.

\subsubsection{Guided Bayesian Optimization (GBO)}
\label{sec:gbo_sampler}
The uninformed exploration of the \ac{rns} sampler is addressed by our second sampler, which has two components to perform informed exploration and active sampling. First, a probability function model is fitted across all the previously explored variables. This feedback allows the model to learn the search space region with large uncertainty that will return a higher objective value. A \ac{gp} model is used to fit the function. Second, an acquisition function that strategically finds the succeeding variables to optimize the objective function. We use the \ac{ucb}~\cite{auer2002using} acquisition function that provides the $\beta$ hyperparameter to control the exploration vs. exploitation trade-off (See \cref{eqn:ucb}). We also include the sampling constraints to restrict the region where the acquisition function looks for the next sampling variables. \cref{algo:bo} shows the operation of the \ac{gbo} sampler, which is discussed below. 

\textit{First}, the \ac{gp} model with properties $\mu$[0] and $\sigma$[0] is initialized for the first k iterations. During these iterations, the scene variables $s_v$ are randomly sampled from their respective distribution range $v_d$. These randomly selected variables are used to generate the scene artifact files, run the simulator and compute the risk score $S_{Risk}$ that is added to the exploration list $\mathcal{E}$ along with the sampled variables. After the k initial iterations, the initialized \ac{gp} model is fitted across all the entries in $\mathcal{E}$ to calculate new posterior distribution $f(x_n)$ with updated $\mu$[$x_n$] and $\sigma$[$x_n$]. 


\textit{Second}, the algorithm uses the sampling constraints $\mathcal{SC}$ to create a smaller search space $\mathcal{B}$ in which the acquisition function will sample the succeeding variables. The updated posterior distribution and the bounded search space created by the sampling constraints are used by an acquisition function to strategically select the following scene variables that will optimize $S_{Risk}$. In this work, we use the \ac{ucb} \cite{auer2002using} as the acquisition function, which is given below. 
 
\begin{equation}
\label{eqn:ucb}
    x_{n+1} = \argmax\limits_{x \in \mathcal{D}}\mu[x_n] + \beta^{1/2} \cdot \sigma[x_n]
\end{equation}

$x_{n+1}$ are the newly selected variables that have the largest \ac{ucb}. $\mu[x_n]$ and $\sigma[x_n]$ are the properties of the updated \ac{gp} model. $\beta$ is a parameter that allows for exploitation and exploration tradeoff. A larger value of $\beta$ allows for higher exploration, and a lower value allows for exploitation. So, $\beta$ needs to be carefully selected for optimizing the two strategies. The newly selected variables are used to run the simulation and compute $S_{Risk}$, which is added to the list $\mathcal{E}$ along with the selected variables. The steps of the algorithm are repeated for a specified number of iterations $t$.  

The sampler has two problems that arise because of using the \ac{gp} model. (1) Cold-start, which requires the model to be trained from scratch each time the sampler is used. This increases the sampling time. To address this, we use the knowledge of randomly sampled scenes from the previous runs to ``warm start" the search process. (2) Scalability, the \ac{gp} model suffers from a cubic time complexity~\cite{liu2020gaussian}, which limits its applicability to a large number of executions. 



\section{Evaluation}
\label{sec:evaluation}

\subsection{\ac{av} Case Study}
\label{sec:casestudy}


Our testbed is an \ac{av} in the CARLA autonomous driving challenge~\cite{carla-challenge} setup, which is needed to navigate an urban town with adverse weather conditions and sensor faults while avoiding collisions with vehicles and pedestrians in its travel path. The experiments with the simulator were performed on a desktop computer with AMD Ryzen Threadripper 16-Core Processor, 4 NVIDIA Titan XP GPUs, and 128 GiB RAM. The details of \ac{av} setup are discussed below. 



\subsubsection{System Model} The system block diagram of the \ac{av} is shown in \cref{fig:sys-model}. It uses a total of $9$ sensors including, three forward-looking cameras, two radars, an \ac{imu}, a \ac{gps}, and a speedometer. It has a \ac{lec} based navigation controller, which is adapted from Chen et al.~\cite{chen2020learning}. This controller uses a navigation planner that takes the waypoint information from the scene artifact file generated by our scene generation approach and divides the distance between the waypoints into smaller position targets. Then, it uses the \ac{gps} and \ac{imu} sensors to get the vehicle's current position and the next position to reach. Next, the position information is fed into a velocity planner to compute the vehicle's desired speed. The desired speed and camera images are fed into a perception \ac{lec}, which predicts the throttle and steering angle errors. These signals are sent to PID controllers to compute the throttle, brake, and steer control signals. Besides the \ac{lec}, there is also an \ac{aebs}, which uses the radar estimated object distance ($r_d$) with the vehicle's current speed to compute a ``safe braking distance" ($b_d$). If $r_d \leq b_d$, a brake alarm is issued. The alarm overrides the throttle signals from the \ac{lec}. Finally, these control signals are sent to the simulator. 

\subsubsection{System Operational Phases} The \ac{av} is designed and operated in the following phases. In the \textit{Design phase}, we design the system model and perform a hazard analysis to identify the operational hazards to the \ac{av}. Next, we identify sensor faults (camera blur and occlusion) that affect the \ac{av}'s performance. In the \textit{Training phase} we use an autopilot controller to collect data for training the perception \ac{lec} using the procedure discussed in~\cite{chen2020learning}. In the \textit{Calibration phase}, we train the detectors and the \ac{resonate} risk estimator. We train the detectors on the training set and the risk estimator on a calibration set, including scenes from the training set and additional scenes with sensor faults and adverse weather conditions, collected using random and grid samplers. Finally, in the \textit{Testing Phase}, the trained \ac{lec} and the safety components are tasked to operate in $250$ scenes generated by our samplers. These scenes included varying the weather conditions (cloud (C), precipitation (P)), time-of-day (T), traffic density (TD), road segments (RS), and sensor faults (F). The distribution ranges of these scene variables and their sampling constraints are listed in \cref{Table:constraints}. \cref{fig:scenes} illustrates the screenshots of a few CARLA scenes generated by our samplers.

\subsubsection{Detectors} 
\label{sec:safety_components}
To detect camera-related faults such as blurred images and occluded images, each of the three cameras is equipped with OpenCV based blur detector and occlusion detector. The blur detector uses the variance of the Laplacian operator~\cite{pech2000diatom} to measure the blur level in the images. A high variance shows that the image is not blurred, and a low variance ($<50$) shows the image is blurred. The occlusion detector is designed to identify large blobs of continuous black image pixels. In our setup, if an image has connected black pixels $>10\%$, then the image is said to be occluded; otherwise, it is not. The blur and occlusion detectors had an F1-score of 97\% and 98\% on the calibration set. 


In addition, we designed a reconstruction-based $\beta$-VAE \ac{ood} detector~\cite{sundar2020out} to identify shifts in the operating scenes. The detector model has four convolutional layers $16/32/64/128$ with (3x3) filters and (2x2) max-pooling followed by four fully connected layers with $2048$, $1000$, and $250$ neurons. It also has a symmetric deconvolutional decoder structure and hyperparameters of $\beta$=1.2 and latent space=$100$. We trained the detector for $150$ epochs using the $6000$ camera images curated for training the \ac{av}. We tested the detector on a test set that included images from the training set and several images with high scene brightness, adverse conditions, and camera faults. For these images, the detector had an F1-score of 95.9\%. At runtime, the reconstruction mean squared error of the detector is used with Inductive Conformal Prediction \cite{shafer2008tutorial} and power martingale \cite{fedorova2012plug} to compute a martingale value. We compared the martingale value with a threshold of $20$, which was empirically selected for \ac{ood} detection. 

\begin{figure*}[t]
 \centering
 \includegraphics[width=0.9\textwidth]{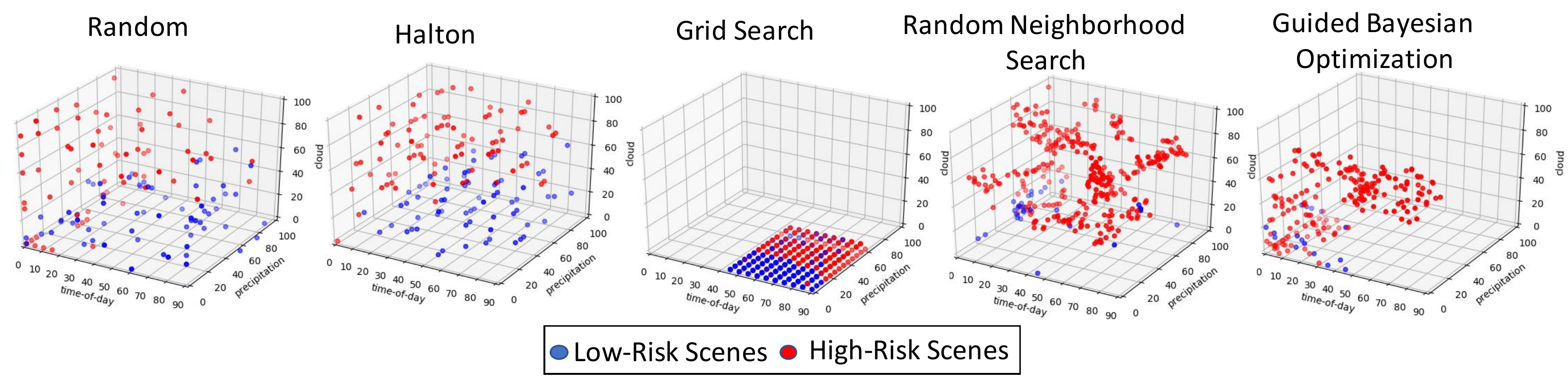}
 \caption{Comparison of the $250$ scenes sampled by the different samplers. While random and Halton samplers only explore the space, grid sampler only exploits. Our samplers balance the exploration vs. exploitation, which is evident from the dense scene clusters in different regions of the search space.}
 \label{fig:sampler_3d}
\vspace{-0.1in}
\end{figure*}

\subsubsection{Risk Score}
\label{sec:risk}
To compute $S_{Risk}$, we first compute the \ac{resonate} score using the \ac{btd} with the hazard of ``roadway obstruction" shown in \cref{fig:btd}. To compute the collision rate $\lambda$, we analyze the conditional relationships of \ac{btd} barriers and threats using the calibration set. The probability function of barriers $B1$ and $B2$ are dependent on the continuous-valued output of the \ac{ood} detector and the binary state of the anomaly detectors, which we capture using the following equation.



\small
\begin{gather}
f_b(x=(B_1,B_2),s) = (1-P(x|s.m.LEC)) \cdot \prod_{d\epsilon S^{B2}}{\frac{P(x|d)}{0.4} \cdot f_{s}} \nonumber \\
P(x|s.m.LEC) = (1+e^{-0.049 \cdot (t.m.LEC-5.754)})^{-1}
\label{eqn:conditional_prob}
\end{gather}
\normalsize

$P(x|s.m.LEC)$ is a sigmoid function used to capture the continuous values of the \ac{ood} detector, the discrete-valued probability of the anomaly detectors that constitutes the state-variables in $S^{B2}$. In our example, $B1$ and $B2$ were affected by the image quality, which we detect using the blur and occlusion detectors whose probabilities are reported in~\cref{fig:btd}. Also, $f_{s}$ is the failure rate of the sensors, which is assumed to be constant for both the camera and the radar sensors on which the barriers are dependent. Further, the probability function of $B3$ is conditionally dependent on the precipitation levels and the operation of the radar as shown in \cref{fig:btd}. These conditional relationships were inferred by clustering the calibration scenes based on the precipitation levels. In a similar approach, we also estimated the frequency of threat occurrence. In this example, the frequency of threats $T1$ and $T2$ depended on the road segments. An intersection had a higher frequency of threats than a side road. Finally, we use these probabilities to compute the dynamic collision rates and the \ac{resonate} score as discussed in \cref{sec:scoring}.  

Next, we compute the infraction score using $IS = 0.7 \cdot I_S + 0.8 \cdot I_R + I_{RD}$. Where $I_S$ is a stop sign infraction, $I_R$ is a red-light infraction, and $I_{RD}$ is the deviation in the route taken by the \ac{av}. This score is added to the \ac{resonate} score to compute the $S_{Risk}$. Finally, we compute the risk threshold $\delta$ to be $0.65$, as discussed in \cref{sec:scoring}. 

\subsection{Baselines and Comparison Metrics} We compare our sampler to state-of-the-art baselines. (1) Random Search is a technique that samples the scene variables uniformly at random from their respective distributions. (2) Grid Search is a technique that exhaustively samples all the combinations of the scene variables in a given grid. (3) Halton Sequence Search~\cite{halton1960efficiency} is a pseudo-random technique that samples the scene variables using co-prime as its bases. We compare these baselines using the following metrics, which align with the sampler objectives discussed in \cref{sec:ps}. 

\textbf{1) Total Risk Scenes (TRS)}: The proportion of the high-risk scenes sampled to the total scenes sampled ($N$).
\begin{equation}
\label{eqn:rsp}
\small
    TRS (\%) = \frac{1}{N}\sum_{i=1}^{N} s_i \quad \textnormal{if} \quad S_{Risk}^i > \delta
\end{equation}
$S_{Risk}^i$ is the risk across the $i^{th}$ sampled scene, and $\delta$ is the risk threshold computed across the calibration set. 



\textbf{2) Diversity ($D$)}: We measure the coverage diversity using two cluster-based quantities. (1) the number of optimal clusters that can be generated from the $N$ sampled scenes, as measured using silhouette score~\cite{rousseeuw1987silhouettes}. (2) the risk variance across each scene across all the clusters. These quantities are combined into computing the diversity score as follows. 

\begin{equation}
\small
\label{eqn:diverse}
    Diversity = \sigma^2\{\frac{1}{len(\mathcal{C})}(\sum_{i=1}^{len(\mathcal{C})} S_{Risk}^i)|\forall \mathcal{C} \in N\}
\end{equation}    

$\mathcal{C}$ is a cluster of scenes generated on the scene variables. $S_{Risk}^i$ is the risk of the $i^{th}$ scene in a cluster. To select the optimal number of clusters, we perform k-means clustering~\cite{wagstaff2001constrained} using the silhouette score as the selection metric.



\textbf{3) Search Time}: The overall time taken by the sampler to sample $N$ scenes and execute them in the simulator.

\subsection{Results}
\label{sec:results}


To compare our samplers against the baselines, we executed each sampler for $250$ iterations. We started the samplers at the same initial condition with time-of-day = $45^{\circ}$ (noon) and a value of zero for the other variables. Also, instead of starting the \ac{gbo} sampler from scratch, we provide it a ``warm start" using the results from the random sampler. Further, we empirically selected the following parameters for controlling the exploration vs. exploitation tradeoff. (1) Number of neighbors $K$ = 6, and threshold $\tau$ = 10 for the \ac{rns} sampler, and (2) $\beta$ = 30 for the \ac{ucb} function of the \ac{gbo} sampler. The simulation operates at a fixed rate of $20$ Frame Per Seconds in these experiments. \cref{fig:sampler_3d} shows the qualitative comparison of the scenes sampled by the samplers. The random and Halton samplers explore the search space, but they cannot exploit it because of their passive search strategy. On the other hand, the grid sampler orderly exploits every point in the grid, limiting its exploration. In comparison, the \ac{rns} and \ac{gbo} samplers perform balanced exploration and exploitation. This is evident from the dense scene clusters at different regions of the search space. Further, we compared the samplers using the comparison metrics. 

\begin{table}[!t]
\centering
\renewcommand{\arraystretch}{1}
\footnotesize
\begin{tabular}{|c|c|c|c|c|c|}
\hline
\multirow{3}{*}{\textbf{Sampler}} & \multirow{3}{*}{\textbf{\begin{tabular}[c]{@{}c@{}}Total \\ Risk \\ Scenes\\ (\%)\end{tabular}}} & \multicolumn{3}{c|}{\textbf{Diversity}}                                                                                                                                                                                         & \multirow{3}{*}{\textbf{\begin{tabular}[c]{@{}c@{}}Search\\ Time \\ (mins)\end{tabular}}} \\ \cline{3-5}
                                  &                                                                                                   & \multicolumn{2}{c|}{\textbf{Cluster Selection}}                                                                                           & \multirow{2}{*}{\textbf{\begin{tabular}[c]{@{}c@{}}Diversity\\ Score\end{tabular}}} &                                                                                           \\ \cline{3-4}
                                  &                                                                                                   & \textbf{\begin{tabular}[c]{@{}c@{}}\# of \\ Clusters\end{tabular}} & \textbf{\begin{tabular}[c]{@{}c@{}}Silhouette \\ score\end{tabular}} &                                                                                     &                                                                                           \\ \hline
\textbf{Random}                   & 66                                                                                                & 3                                                                  & 0.34                                                                 & 0.02                                                                                & 323                                                                                       \\ \hline
\textbf{Halton}                   & 71                                                                                                & 2                                                                  & 0.27                                                                 & 0.07                                                                                & 315                                                                                       \\ \hline
\textbf{Grid}                     & 56                                                                                                & 2                                                                  & 0.71                                                                 & 0.135                                                                                &  309                                                                                      \\ \hline
\textbf{RNS}                      & 83                                                                                                & 6                                                                  & 0.56                                                                 & 0.193                                                                                & 332                                                                                       \\ \hline
\textbf{GBO}                      & 92                                                                                                & 4                                                                  & 0.62                                                                 & 0.632                                                                                   & 897                                                                                       \\ \hline
\end{tabular}
\caption{Comparing the sampling approaches using the comparison metrics.}
\label{Table:comparison}
\end{table}

\begin{figure}[!t]
\centering
\begin{tikzpicture}
\footnotesize

\begin{groupplot}[group style = {group size = 1 by 2, horizontal sep = 55pt}, width = 6.0cm, height = 5.0cm]
        \nextgroupplot[ 
            height=3cm, width=7cm,
        bar width=0.2cm,
    ybar,
    enlargelimits=0.15,
    x grid style={white!69.0196078431373!black},
xmajorgrids,
y grid style={white!69.0196078431373!black},
ymajorgrids,
    legend style={at={(0.5,-0.4),font=\tiny},
      anchor=north,legend columns=-1},
    ylabel style={align=center}, ylabel= OOD\\ Scenes (\%),
    symbolic x coords={Random,Halton,Grid,RNS,GBO},
    xtick=data,
    xtick pos=left,
    ymin=0, ymax=100,
    ytick={0,25,50,75,100},
ytick pos=left,
    nodes near coords align={vertical},
    ]
\addplot [blue!20!black,fill=blue!80!white]coordinates 
{(Random,54.4)(Halton,60)(Grid,39)(RNS,72)(GBO,79.2)};
        \nextgroupplot[height=3cm, width=7cm,
        bar width=0.2cm,
    ybar, area legend,
    enlargelimits=0.15,
    x grid style={white!69.0196078431373!black},
xmajorgrids,
y grid style={white!69.0196078431373!black},
ymajorgrids,
    legend style={at={(-0.2,-0.4),font=\tiny},
      anchor=north,legend columns=-1},
   ylabel style={align=center}, ylabel= Scenes \\ with \\ Infractions (\%),
    symbolic x coords={Random,Halton,Grid,RNS,GBO},
    xtick=data,
    xtick pos=left,
    ymin=0, ymax=100,
    ytick={0,25,50,75,100},
ytick pos=left,
    nodes near coords align={vertical},
    ]
            \addplot[red!20!black,fill=red!80!white] coordinates 
            {(Random,14.8)(Halton,22.8)(Grid,19.4)(RNS,29)(GBO,26.5)};

    \end{groupplot}

\end{tikzpicture}
\caption{(Top) Total \ac{ood} scenes and (Bottom) Total scenes with infractions sampled among the total $250$ scenes. Our samplers sampled a higher percent of \ac{ood} scenes and scenes with infractions.}
\label{fig:infractions}
\vspace{-0.15in}
\end{figure}
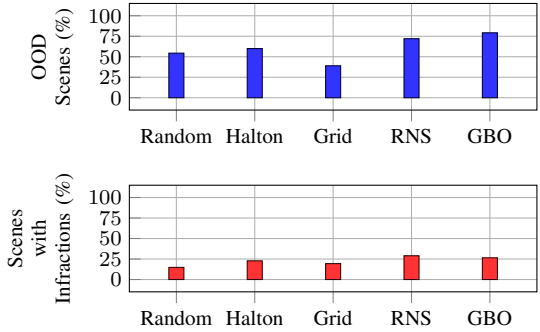


\textbf{Total Risk Scenes}: \cref{Table:comparison} lists the total risk scenes generated by different samplers. The random, Halton, and grid samplers sampled $66$\%, $71$\%, and $56$\% of high-risk scenes, respectively. In comparison, the \ac{rns} sampler sampled 83\% and \ac{gbo} sampled 92\% high-risk scenes among the $250$ scenes that were sampled. Also, \cref{fig:infractions} illustrates the number of \ac{ood} scenes and scenes with infractions that were generated by the samplers in $250$ iterations. As seen, the \ac{gbo} and the \ac{rns} samplers generated higher \ac{ood} scenes compared to the baselines. Also, the \ac{rns} sampler sampled the highest scenes with infractions followed by \ac{gbo} and Halton samplers. The actual infractions are low even when the risk is high for the following reasons. (1) the \ac{lec} controller was sufficiently trained to avoid \ac{av} infractions, (2) the \ac{aebs} worked sufficiently well to stop the \ac{av}.

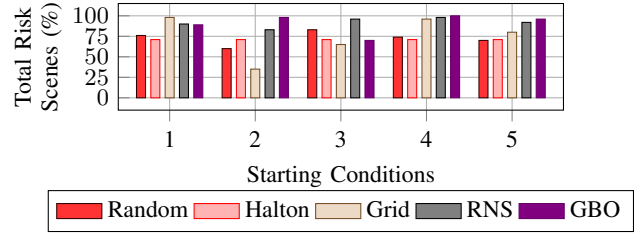
\begin{figure}[!t]
\centering
\begin{tikzpicture}
\begin{axis}[
height=3cm, width=7.5cm,
        bar width=0.12cm,
        ylabel style={yshift=-0.1cm},
    ybar=2pt, area legend,
    x grid style={white!69.0196078431373!black},
xmajorgrids,
y grid style={white!69.0196078431373!black},
ymajorgrids,
    enlargelimits=0.15,
    label style={font=\small},
    tick label style={font=\small},
    legend style={at={(0.5,-0.75),font=\small},
      anchor=north,legend columns=-1},
    xlabel={Starting Conditions},
    ylabel style={align=center}, ylabel= Total Risk \\ Scenes (\%),
    symbolic x coords={1,2,3,4,5},
    xtick=data,
    xtick pos=left,
    ymin=0, ymax=100,
    ytick={0,25,50,75,100},
    ytick pos=left,
    nodes near coords align={vertical},
    ]
\addplot[red!20!black,fill=red!80!white] coordinates {(1,76)(2,60)(3,83)(4,74)(5,70)};
\addplot coordinates {(1,71)(2,71)(3,71)(4,71)(5,71)};
\addplot coordinates {(1,98)(2,35)(3,65)(4,96)(5,80)};
\addplot coordinates {(1,90)(2,83)(3,96)(4,98)(5,92)};
\addplot coordinates {(1,89)(2,98)(3,70)(4,100)(5,96)};
 \legend{Random, Halton, Grid, RNS, GBO}            
\end{axis}
\end{tikzpicture}
\caption{Total high risk scenes sampled by samplers starting from different initial conditions. Generally, our samplers sampled higher percent of risky scenes across all the starting conditions.}
\label{fig:initial}
\vspace{-0.15in}
\end{figure}


We also started the simulator at $5$ different initial conditions. (1) T = $0^{\circ}$ (dusk), (2) T = $45^{\circ}$ (noon), (3) T = $90^{\circ}$ (noon), (4) T = $45^{\circ}$ (noon), P = 50\%, C = 50\%, and (5) T = $45^{\circ}$, P = 0\%, C = 50\%, and the other variables were set to zero. Each sampler was executed for $50$ iterations from these initial conditions. \cref{fig:initial} shows the high-risk scenes generated by the different samplers. The \ac{rns} and \ac{gbo} samplers generated high-risk scenes across the different starts as compared to the baselines. Among the baselines, the grid search generated higher high-risk scenes than the other two samplers. To note, different starting conditions did not affect the Halton sampler.

\textbf{Diversity}: To compute diversity, we classified the sampled scenes into clusters by applying k-means clustering analysis along with silhouette score. As seen in \cref{Table:comparison}, the scenes generated by the \ac{rns} and \ac{gbo} samplers could be classified into a higher number of clusters ($6$ and $4$) as compared to the Halton, grid, and random samplers that could only be classified into $2$, $2$, and $3$ clusters, respectively. We used these clusters to compute the diversity score in \cref{eqn:diverse}. As seen, the \ac{gbo} and \ac{rns} samplers had a higher diversity score of $0.63$ and $0.19$ as compared to the Halton, random, and grid samplers, which had scores of $0.07$, $0.02$, and $0.13$, respectively. The \ac{gbo} and \ac{rns} samplers had higher coverage diversity because they sampled a higher number of scenes around a scene that was previously found to be of high risk to the \ac{av}.




\begin{figure}[!t]
\begin{tikzpicture}
\pgfmathsetlengthmacro\MajorTickLength{
      \pgfkeysvalueof{/pgfplots/major tick length} * 0.5
    }
\begin{groupplot}[group style={group size=3 by 1,horizontal sep = 0.8 cm, vertical sep = 1.2cm}]
\nextgroupplot[
title=\textbf{Random},
title style={yshift=-1ex,,font=\small},
      width=0.42\columnwidth,
      height=0.2\textwidth,
      font=\footnotesize,
      grid=major,
      xmin=0, xmax=4,
      xtick={0,1,2,3,4},
      yticklabels={{C1}, {C2},{C3}},
       ytick={1,2,3},
       ResourceOfferStyle/.style={boxplot={box extend=0.9,}, green, solid, fill=green!80, mark=x},
      RegisterStyle/.style={boxplot={box extend=0.5,}, blue, solid, fill=blue!40, mark=x},
      OfferMatchTimeStyle/.style={boxplot={box extend=0.5,}, red, solid, fill=red!80, mark=x},
      ResultTimeStyle/.style={boxplot={box extend=0.5,}, gray, solid, fill=gray!80, mark=x},
      ExecutionTimeStyle/.style={boxplot={box extend=0.9,}, orange, solid, fill=orange!80, mark=x},
      xlabel={$S_{Risk}$}
    ]
    \addplot+[RegisterStyle, boxplot={draw position=1}] table [col sep=comma, y=c1] {result-csv/random_risk1.csv};
     \addplot+[OfferMatchTimeStyle, boxplot={draw position=2}] table [col sep=comma, y=c2] {result-csv/random_risk1.csv}; 
    \addplot+[ResourceOfferStyle, boxplot={draw position=3}] table [col sep=comma, y=c3] {result-csv/random_risk1.csv}; 

\nextgroupplot[
title=\textbf{Halton},
title style={yshift=-1ex,,font=\small},
      width=0.42\columnwidth,
      height=0.2\textwidth,
      font=\footnotesize,
      grid=major,
      xmin=0, xmax=4,
      xtick={0,1,2,3,4},
      yticklabels=\empty,
      yticklabels={{C1}, {C2}},
       ytick={1,2},
       ResourceOfferStyle/.style={boxplot={box extend=0.9,}, green, solid, fill=green!80, mark=x},
      RegisterStyle/.style={boxplot={box extend=0.5,}, blue, solid, fill=blue!40, mark=x},
      OfferMatchTimeStyle/.style={boxplot={box extend=0.5,}, red, solid, fill=red!80, mark=x},
      ResultTimeStyle/.style={boxplot={box extend=0.5,}, gray, solid, fill=gray!80, mark=x},
      ExecutionTimeStyle/.style={boxplot={box extend=0.9,}, orange, solid, fill=orange!80, mark=x},
      MyExecutionTimeStyle/.style={boxplot={box extend=0.9,}, black, solid, fill=black!80, mark=x},
      xlabel={$S_{Risk}$}
    ]
    \addplot+[RegisterStyle, boxplot={draw position=1}] table [col sep=comma, y=c1] {result-csv/halton_risk.csv};
     \addplot+[OfferMatchTimeStyle, boxplot={draw position=2}] table [col sep=comma, y=c2] {result-csv/halton_risk.csv};
     
\nextgroupplot[
title=\textbf{Grid},
title style={yshift=-1ex,,font=\small},
      width=0.42\columnwidth,
      height=0.2\textwidth,
      font=\footnotesize,
      grid=major,
      xmin=0, xmax=4,
      xtick={0,1,2,3,4},
      yticklabels={{C1}, {C2}},
       ytick={1,2},
       ResourceOfferStyle/.style={boxplot={box extend=0.9,}, green, solid, fill=green!80, mark=x},
      RegisterStyle/.style={boxplot={box extend=0.5,}, blue, solid, fill=blue!40, mark=x},
      OfferMatchTimeStyle/.style={boxplot={box extend=0.5,}, red, solid, fill=red!80, mark=x},
      ResultTimeStyle/.style={boxplot={box extend=0.5,}, gray, solid, fill=gray!80, mark=x},
      ExecutionTimeStyle/.style={boxplot={box extend=0.9,}, orange, solid, fill=orange!80, mark=x},
      xlabel={$S_{Risk}$}
    ]
    \addplot+[RegisterStyle, boxplot={draw position=1}] table [col sep=comma, y=c1] {result-csv/grid_risk.csv};
     \addplot+[OfferMatchTimeStyle, boxplot={draw position=2}] table [col sep=comma, y=c2] {result-csv/grid_risk.csv};

\end{groupplot}
\end{tikzpicture}
 \vspace{-0.1cm} 
 \begin{tikzpicture}
\pgfmathsetlengthmacro\MajorTickLength{
      \pgfkeysvalueof{/pgfplots/major tick length} * 0.5
    }
\begin{groupplot}[group style={group size=3 by 1,horizontal sep = 0.8 cm, vertical sep = 1.2cm}]
\nextgroupplot[
title=\textbf{RNS},
title style={yshift=-1ex,,font=\small},
      width=0.5\columnwidth,
      height=0.22\textwidth,
      font=\footnotesize,
      grid=major,
      xmin=0, xmax=4,
      xtick={0,1,2,3,4},
      yticklabels=\empty,
      yticklabels={{C1}, {C2},{C3},{C4},{C5},{C6}},
       ytick={1,2,3,4,5,6},
       ResourceOfferStyle/.style={boxplot={box extend=0.9,}, green, solid, fill=green!80, mark=x},
      RegisterStyle/.style={boxplot={box extend=0.5,}, blue, solid, fill=blue!40, mark=x},
      OfferMatchTimeStyle/.style={boxplot={box extend=0.5,}, red, solid, fill=red!80, mark=x},
      ResultTimeStyle/.style={boxplot={box extend=0.5,}, gray, solid, fill=gray!80, mark=x},
      ExecutionTimeStyle/.style={boxplot={box extend=0.9,}, orange, solid, fill=orange!80, mark=x},
      MyExecutionTimeStyle/.style={boxplot={box extend=0.9,}, black, solid, fill=black!80, mark=x},
      xlabel={$S_{Risk}$}
    ]
    \addplot+[RegisterStyle, boxplot={draw position=1}] table [col sep=comma, y=c1] {result-csv/rns_risk.csv};
     \addplot+[OfferMatchTimeStyle, boxplot={draw position=2}] table [col sep=comma, y=c2] {result-csv/rns_risk.csv};
    \addplot+[ResourceOfferStyle, boxplot={draw position=3}] table [col sep=comma, y=c3] {result-csv/rns_risk.csv}; 
    \addplot+[ResultTimeStyle, boxplot={draw position=4}] table [col sep=comma, y=c4] {result-csv/rns_risk.csv}; 
    \addplot+[ExecutionTimeStyle, boxplot={draw position=5}] table [col sep=comma, y=c5] {result-csv/rns_risk.csv};
    \addplot+[MyExecutionTimeStyle, boxplot={draw position=6}] table [col sep=comma, y=c6] {result-csv/rns_risk.csv};
     
\nextgroupplot[
title=\textbf{GBO},
title style={yshift=-1ex,,font=\small},
      width=0.5\columnwidth,
      height=0.22\textwidth,
      font=\footnotesize,
      grid=major,
      xmin=0, xmax=4,
      xtick={0,1,2,3,4},
      yticklabels={{C1}, {C2},{C3},{C4}},
       ytick={1,2,3,4},
       ResourceOfferStyle/.style={boxplot={box extend=0.9,}, green, solid, fill=green!80, mark=x},
      RegisterStyle/.style={boxplot={box extend=0.5,}, blue, solid, fill=blue!40, mark=x},
      OfferMatchTimeStyle/.style={boxplot={box extend=0.5,}, red, solid, fill=red!80, mark=x},
      ResultTimeStyle/.style={boxplot={box extend=0.5,}, gray, solid, fill=gray!80, mark=x},
      ExecutionTimeStyle/.style={boxplot={box extend=0.9,}, orange, solid, fill=orange!80, mark=x},
      xlabel={$S_{Risk}$}
    ]
    \addplot+[RegisterStyle, boxplot={draw position=1}] table [col sep=comma, y=c1] {result-csv/gbo_risk.csv};
     \addplot+[OfferMatchTimeStyle, boxplot={draw position=2}] table [col sep=comma, y=c2] {result-csv/gbo_risk.csv};
    \addplot+[ResourceOfferStyle, boxplot={draw position=3}] table [col sep=comma, y=c3] {result-csv/gbo_risk.csv}; 
    \addplot+[ResultTimeStyle, boxplot={draw position=4}] table [col sep=comma, y=c4] {result-csv/gbo_risk.csv}; 
    
\end{groupplot}
\end{tikzpicture}
\footnotesize
\centering
\setlength{\abovecaptionskip}{2pt}
\caption{The variance of $S_{Risk}$ across scene clusters generated by different samplers. Our samplers generate more clusters with a high variance of $S_{Risk}$. Y-axis shows the number of clusters, and the Y-axis shows the $S_{Risk}$ score.}
\label{fig:cluster}
\vspace{-0.1in}
\end{figure}
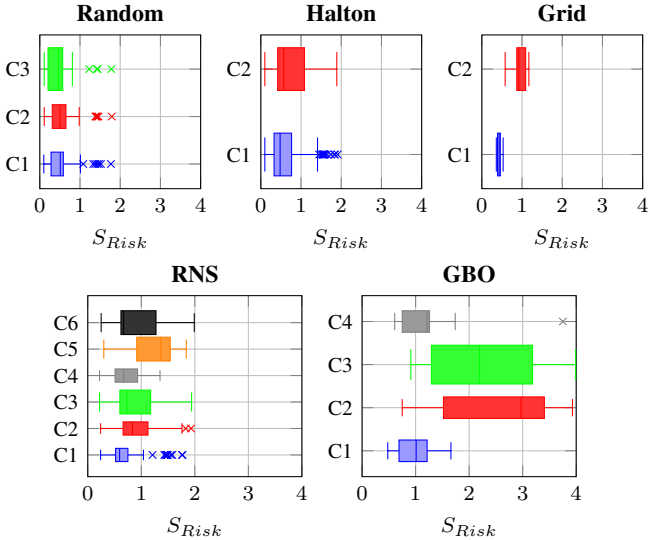

\textbf{Search Times}: We report the search times of the samplers in \cref{Table:comparison}. The random, Halton, grid, and \ac{rns} samplers have comparable search times of $323$, $315$, $309$, and $332$ minutes, respectively. However, the \ac{gbo} sampler required the highest search time of $897$ minutes because of its scalability issue discussed in \cref{sec:gbo_sampler}. Further, for a direct comparison, we counted the number of scenes sampled by each sampler within a fixed budget of $60$ minutes. The random, Halton, grid and 
\ac{rns} samplers sampled $47$, $51$, $49$, and $46$ scenes in $60$ minutes. The \ac{gbo} sampler only sampled $19$ scenes. 



\subsection{Discussion}
\label{sec:discussion}
The key takeaways from the experiments are. \underline{First}, we observe that the \ac{rns} and \ac{gbo} samplers outperform the baselines by generating a higher percentage of high-risk scenes (first objective). \underline{Second}, we observe that the scene generated by the \ac{rns} and \ac{gbo} samplers have a higher diversity score as compared to the baselines (second objective). \underline{Third}, the \ac{gbo} sampler is the most effective in generating high-risk scenes, and its informed exploration results in diverse search. However, the use of \ac{gp} model increases the search time, which we plan to address in the future using a scalable \ac{gp} model. In comparison, the \ac{rns} sampler is efficient and requires lower search times to generate high-risk scenes. 

\begin{figure}[t]
\centering
 \includegraphics[width=\columnwidth]{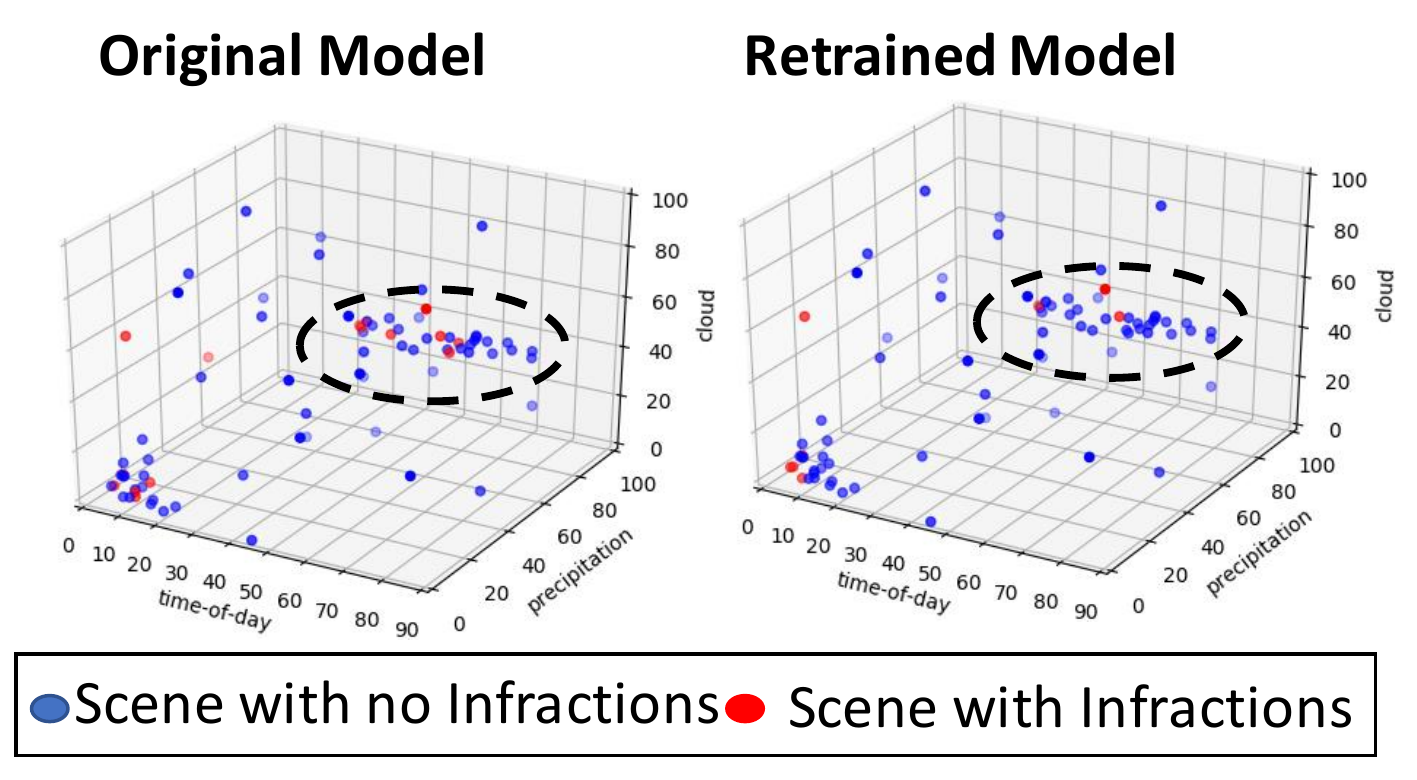}
\caption{Performance of $M_{retrain}$ and $M_{orig}$ models. $M_{retrain}$ model had fewer infractions, especially in the scenes from the region highlighted.}
\label{fig:retrain}
\vspace{-0.1in}
\end{figure}

\underline{Lastly}, the high-risk scenes identified by our samplers can be used to improve the system's \acp{lec}. To illustrate, the \ac{av} system in our case study was susceptible to scenes from dusk with high precipitation ($T<45$ and $P>50$) as identified by our samplers. So, we trained a new \ac{lec} controller model $M_{retrain}$ on a dataset that contained both the original train scenes and $50$ high-risk scenes sampled by the \ac{rns} sampler. We trained the model for $50$ epochs and saved several model weights, and utilized the weights that optimized the \ac{av}'s driving proficiency. We then compared the performance of the original model $M_{orig}$ (model trained on original training scenes) and the retrained model $M_{retrain}$ in terms of the actual system infractions, which included collision, running a red light or stop sign, and route deviation. The comparison is shown in \cref{fig:retrain}, and for this, we curated $80$ operational scenes that included random scenes and scenes where the system previously failed. We observed that $M_{retrain}$ had only $6$ scenes with infractions as compared to $11$ scenes of $M_{orig}$. Retraining the model on the high-risk scenes reduced the system infractions, especially those from the highlighted region where it previously had higher infractions. Additionally, we observed that $M_{retrain}$ had improved driving proficiency with fewer route deviations for the high-risk scenes. However, both models performed several red-light infractions in general, and retraining did not address the problem.


\section{Related Work}
\label{sec:rw}
Tools like Dakota~\cite{dalbey2021dakota} with sampling approaches like incremental sampling, importance sampling, and adaptive sampling have been long available for uncertainty quantification in engineering design. Such samplers have recently gained interest for sampling simulation-based scenes in the \ac{av} domain. Several domain-specific \acp{sdl} like Scenic~\cite{fremont2019scenic} and MSDL~\cite{msdl} have been designed for specifying scenarios of complex traffic conditions and operating conditions to identify counterexamples that affect the system's safety. These languages use probabilistic samplers that sample scenes across the entire search space formed by the combination of scene variables. For example, Grid search is a popular passive sampler, which takes a single risky scene as the starting condition and generates similar risky scenes around it. However, manual tuning makes the grid sampler labor-intensive and time-consuming. For accelerating the search process, an alternate approach~\cite{zhao2017accelerated} uses importance sampling to identify important scene variables and sample scenes using them. Worst-Case Scenario Evaluation~\cite{ma1999worst} is another approach that uses model-based optimization to identify the weakness of the system and uses this information to generate worst-case disturbances. VERIFAI~\cite{dreossi2019verifai} software toolkit has several passive samplers like random search, Halton search~\cite{halton1960efficiency}, and active samplers like cross-entropy optimization and \acl{bo}. However, these samplers do not effectively balance the exploration vs. exploitation trade-off. Viswanadha, Kesav, \emph{et al.}~\cite{viswanadha2021parallel} recently proposed a multi-armed bandit sampler that balances the two strategies.


Also, recently, there has been growing interest in using generative models for generating risky scenes~\cite{ding2018new,vardhan2021rare,o2018scalable,ding2020learning}. In this approach, scenes are randomly sampled from a distribution learned by a generative model instead of sampling from the entire search space (as performed by the samplers discussed above). For example, Ding, Wenhao, \emph{et al.}~\cite{ding2020learning} use an autoregressive model to factorize the scene variables into conditional probability distributions, which are then sampled to generate risky traffic scenes. In a prior work~\cite{ding2018new}, they have also used a \acl{vae} to project high-dimensional traffic information into a lower-dimensional latent space, which is sampled to generate critical traffic scenes. Vardhan \emph{et al.}~\cite{vardhan2021rare} uses a Gaussian Mixture Model to find the corner case scenes in the training set that is not well learned by the system. Another approach~\cite{o2018scalable} uses generative adversarial imitation learning to perform adaptive importance-sampling to learn rare events from an underlying data distribution.

\section{Conclusion}
\label{sec:conclusion}
We presented a scene generation approach that integrates a \acl{sdl} with two samplers and a risk-based scoring function for generating high-risk scenes. \acl{rns} and \acl{gbo} are the proposed active samplers that perform constraint-based sampling and balance the exploration vs. exploitation to guide them towards sampling clusters of high-risk scenes. We applied these samplers to an \ac{av} case study. Our evaluations show that the proposed samplers could sample a higher number of high-risk scenes that could be clustered into a higher number of clusters than the conventional baselines.


\textbf{Acknowledgement}: This work was supported by the DARPA Assured Autonomy project and Air Force Research Laboratory. Any opinions, findings, and conclusions or recommendations expressed in this material are those of the author(s) and do not necessarily reflect the views of DARPA or AFRL.




\bibliographystyle{IEEEtran}
\bibliography{main.bib}

\begin{thebibliography}{10}
\providecommand{\url}[1]{#1}
\csname url@samestyle\endcsname
\providecommand{\newblock}{\relax}
\providecommand{\bibinfo}[2]{#2}
\providecommand{\BIBentrySTDinterwordspacing}{\spaceskip=0pt\relax}
\providecommand{\BIBentryALTinterwordstretchfactor}{4}
\providecommand{\BIBentryALTinterwordspacing}{\spaceskip=\fontdimen2\font plus
\BIBentryALTinterwordstretchfactor\fontdimen3\font minus
  \fontdimen4\font\relax}
\providecommand{\BIBforeignlanguage}[2]{{%
\expandafter\ifx\csname l@#1\endcsname\relax
\typeout{** WARNING: IEEEtran.bst: No hyphenation pattern has been}%
\typeout{** loaded for the language `#1'. Using the pattern for}%
\typeout{** the default language instead.}%
\else
\language=\csname l@#1\endcsname
\fi
#2}}
\providecommand{\BIBdecl}{\relax}
\BIBdecl

\bibitem{bishop2000methodology}
P.~Bishop and R.~Bloomfield, ``A methodology for safety case development,'' in
  \emph{Safety and Reliability}.\hskip 1em plus 0.5em minus 0.4em\relax Taylor
  \& Francis, 2000.

\bibitem{denney2015dynamic}
E.~Denney, G.~Pai, and I.~Habli, ``Dynamic safety cases for through-life safety
  assurance,'' in \emph{IEEE/ACM 37th IEEE International Conference on Software
  Engineering}, vol.~2.\hskip 1em plus 0.5em minus 0.4em\relax IEEE, 2015, pp.
  587--590.

\bibitem{sundar2020out}
V.~K. Sundar, S.~Ramakrishna, Z.~Rahiminasab, A.~Easwaran, and A.~Dubey,
  ``Out-of-distribution detection in multi-label datasets using latent space of
  $\beta$-vae,'' \emph{arXiv preprint arXiv:2003.08740}, 2020.

\bibitem{ramakrishna2021efficient}
S.~Ramakrishna, Z.~Rahiminasab, G.~Karsai, A.~Easwaran, and A.~Dubey,
  ``Efficient out-of-distribution detection using latent space of $\beta$-vae
  for cyber-physical systems,'' \emph{arXiv preprint arXiv:2108.11800}, 2021.

\bibitem{kuhn2020introspective}
C.~B. Kuhn, M.~Hofbauer, G.~Petrovic, and E.~Steinbach, ``Introspective failure
  prediction for autonomous driving using late fusion of state and camera
  information,'' \emph{IEEE Transactions on Intelligent Transportation
  Systems}, 2020.

\bibitem{kuhn2021trajectory}
------, ``Trajectory-based failure prediction for autonomous driving,'' in
  \emph{2021 IEEE Intelligent Vehicles Symposium (IV)}.\hskip 1em plus 0.5em
  minus 0.4em\relax IEEE, 2021, pp. 980--986.

\bibitem{hartsell2021resonate}
C.~Hartsell, S.~Ramakrishna, A.~Dubey, D.~Stojcsics, N.~Mahadevan, and
  G.~Karsai, ``Resonate: A runtime risk assessment framework for autonomous
  systems,'' in \emph{International Symposium on Software Engineering for
  Adaptive and Self-Managing Systems (SEAMS)}, may 2021.

\bibitem{ma1999worst}
W.-H. Ma and H.~Peng, ``A worst-case evaluation method for dynamic systems,''
  1999.

\bibitem{ding2020learning}
W.~Ding, B.~Chen, M.~Xu, and D.~Zhao, ``Learning to collide: An adaptive
  safety-critical scenarios generating method,'' in \emph{2020 IEEE/RSJ
  International Conference on Intelligent Robots and Systems (IROS)}.\hskip 1em
  plus 0.5em minus 0.4em\relax IEEE, 2020, pp. 2243--2250.

\bibitem{khan2019striking}
S.~Khan, M.~Hayat, S.~W. Zamir, J.~Shen, and L.~Shao, ``Striking the right
  balance with uncertainty,'' in \emph{Proceedings of the IEEE/CVF Conference
  on Computer Vision and Pattern Recognition}, 2019, pp. 103--112.

\bibitem{fremont2019scenic}
D.~J. Fremont, T.~Dreossi, S.~Ghosh, X.~Yue, A.~L. Sangiovanni-Vincentelli, and
  S.~A. Seshia, ``Scenic: A language for scenario specification and scene
  generation,'' in \emph{Proceedings of the 40th annual ACM SIGPLAN conference
  on Programming Language Design and Implementation (PLDI)}, June 2019.

\bibitem{dalbey2021dakota}
K.~Dalbey, M.~Eldred, G.~Geraci, J.~Jakeman, K.~Maupin, J.~A. Monschke,
  D.~Seidl, A.~Tran, F.~Menhorn, and X.~Zeng, ``Dakota a multilevel parallel
  object-oriented framework for design optimization parameter estimation
  uncertainty quantification and sensitivity analysis: Version 6.14 theory
  manual.'' Sandia National Lab.(SNL-NM), Albuquerque, NM (United States),
  Tech. Rep., 2021.

\bibitem{zhao2017accelerated}
D.~Zhao, X.~Huang, H.~Peng, H.~Lam, and D.~J. LeBlanc, ``Accelerated evaluation
  of automated vehicles in car-following maneuvers,'' \emph{IEEE Transactions
  on Intelligent Transportation Systems}, 2017.

\bibitem{dreossi2019verifai}
T.~Dreossi, D.~J. Fremont, S.~Ghosh, E.~Kim, H.~Ravanbakhsh,
  M.~Vazquez-Chanlatte, and S.~A. Seshia, ``Verifai: A toolkit for the formal
  design and analysis of artificial intelligence-based systems,'' in
  \emph{International Conference on Computer Aided Verification}.\hskip 1em
  plus 0.5em minus 0.4em\relax Springer, 2019, pp. 432--442.

\bibitem{viswanadha2021parallel}
K.~Viswanadha, E.~Kim, F.~Indaheng, D.~J. Fremont, and S.~A. Seshia, ``Parallel
  and multi-objective falsification with scenic and verifai,'' in
  \emph{International Conference on Runtime Verification}.\hskip 1em plus 0.5em
  minus 0.4em\relax Springer, 2021.

\bibitem{msdl}
\BIBentryALTinterwordspacing
O.~foretellix, ``Open m-sdl.'' [Online]. Available:
  \url{https://www.foretellix.com/open-language/}
\BIBentrySTDinterwordspacing

\bibitem{dosovitskiy2017carla}
A.~Dosovitskiy, G.~Ros, F.~Codevilla, A.~Lopez, and V.~Koltun, ``Carla: An open
  urban driving simulator,'' \emph{arXiv:1711.03938}, 2017.

\bibitem{jerebic2021novel}
J.~Jerebic, M.~Mernik, S.-H. Liu, M.~Ravber, M.~Baketari{\'c}, L.~Mernik, and
  M.~{\v{C}}repin{\v{s}}ek, ``A novel direct measure of exploration and
  exploitation based on attraction basins,'' \emph{Expert Systems with
  Applications}, vol. 167, p. 114353, 2021.

\bibitem{halton1960efficiency}
J.~H. Halton, ``On the efficiency of certain quasi-random sequences of points
  in evaluating multi-dimensional integrals,'' \emph{Numerische Mathematik},
  vol.~2, no.~1, pp. 84--90, 1960.

\bibitem{carla-challenge}
\BIBentryALTinterwordspacing
O.~CARLA, ``Carla leaderboard challenge,'' 2020. [Online]. Available:
  \url{https://leaderboard.carla.org/}
\BIBentrySTDinterwordspacing

\bibitem{chen2020learning}
D.~Chen, B.~Zhou, V.~Koltun, and P.~Kr{\"a}henb{\"u}hl, ``Learning by
  cheating,'' in \emph{Conference on Robot Learning}.\hskip 1em plus 0.5em
  minus 0.4em\relax PMLR, 2020, pp. 66--75.

\bibitem{cai2020real}
\BIBentryALTinterwordspacing
F.~Cai and X.~Koutsoukos, ``Real-time out-of-distribution detection in
  learning-enabled cyber-physical systems,'' in \emph{2020 ACM/IEEE 11th
  International Conference on Cyber-Physical Systems (ICCPS)}.\hskip 1em plus
  0.5em minus 0.4em\relax Los Alamitos, CA, USA: IEEE Computer Society, 2020.
  [Online]. Available:
  \url{https://doi.ieeecomputersociety.org/10.1109/ICCPS48487.2020.00024}
\BIBentrySTDinterwordspacing

\bibitem{ferdous2013analyzing}
R.~Ferdous, F.~Khan, R.~Sadiq, P.~Amyotte, and B.~Veitch, ``Analyzing system
  safety and risks under uncertainty using a bow-tie diagram: An innovative
  approach,'' \emph{Process Safety and Environmental Protection}, vol.~91, no.
  1-2, pp. 1--18, 2013.

\bibitem{rousseeuw1987silhouettes}
P.~J. Rousseeuw, ``Silhouettes: a graphical aid to the interpretation and
  validation of cluster analysis,'' \emph{Journal of computational and applied
  mathematics}, vol.~20, pp. 53--65, 1987.

\bibitem{dejanovic2017textx}
I.~Dejanovi{\'c}, R.~Vaderna, G.~Milosavljevi{\'c}, and {\v{Z}}.~Vukovi{\'c},
  ``Textx: a python tool for domain-specific languages implementation,''
  \emph{Knowledge-Based Systems}, vol. 115, pp. 1--4, 2017.

\bibitem{friedman1975algorithm}
J.~H. Friedman, J.~L. Bentley, and R.~A. Finkel, \emph{An algorithm for finding
  best matches in logarithmic time}.\hskip 1em plus 0.5em minus 0.4em\relax
  Department of Computer Science, Stanford University, 1975.

\bibitem{auer2002using}
P.~Auer, ``Using confidence bounds for exploitation-exploration trade-offs,''
  \emph{Journal of Machine Learning Research}, vol.~3, no. Nov, 2002.

\bibitem{liu2020gaussian}
H.~Liu, Y.-S. Ong, X.~Shen, and J.~Cai, ``When gaussian process meets big data:
  A review of scalable gps,'' \emph{IEEE transactions on neural networks and
  learning systems}, vol.~31, no.~11, pp. 4405--4423, 2020.

\bibitem{pech2000diatom}
J.~L. Pech-Pacheco, G.~Crist{\'o}bal, J.~Chamorro-Martinez, and
  J.~Fern{\'a}ndez-Valdivia, ``Diatom autofocusing in brightfield microscopy: a
  comparative study,'' in \emph{Proceedings 15th International Conference on
  Pattern Recognition. ICPR-2000}, vol.~3.\hskip 1em plus 0.5em minus
  0.4em\relax IEEE, 2000, pp. 314--317.

\bibitem{shafer2008tutorial}
G.~Shafer and V.~Vovk, ``A tutorial on conformal prediction,'' \emph{Journal of
  Machine Learning Research}, vol.~9, no. Mar, pp. 371--421, 2008.

\bibitem{fedorova2012plug}
V.~Fedorova, A.~Gammerman, I.~Nouretdinov, and V.~Vovk, ``Plug-in martingales
  for testing exchangeability on-line,'' \emph{arXiv preprint arXiv:1204.3251},
  2012.

\bibitem{wagstaff2001constrained}
K.~Wagstaff, C.~Cardie, S.~Rogers, S.~Schr{\"o}dl \emph{et~al.}, ``Constrained
  k-means clustering with background knowledge,'' in \emph{Icml}, vol.~1, 2001.

\bibitem{ding2018new}
W.~Ding, W.~Wang, and D.~Zhao, ``A new multi-vehicle trajectory generator to
  simulate vehicle-to-vehicle encounters,'' \emph{arXiv preprint
  arXiv:1809.05680}, 2018.

\bibitem{vardhan2021rare}
H.~Vardhan and J.~Sztipanovits, ``Rare event failure test case generation in
  learning-enabled-controllers,'' in \emph{2021 6th International Conference on
  Machine Learning Technologies}, 2021, pp. 34--40.

\bibitem{o2018scalable}
M.~O'Kelly, A.~Sinha, H.~Namkoong, J.~Duchi, and R.~Tedrake, ``Scalable
  end-to-end autonomous vehicle testing via rare-event simulation,''
  \emph{arXiv preprint arXiv:1811.00145}, 2018.

\end{thebibliography}

\end{document}